\documentclass{article}

\usepackage{arxiv}

\usepackage[utf8]{inputenc} 
\usepackage[T1]{fontenc}    
\usepackage{hyperref}       
\usepackage{url}            
\usepackage{booktabs}       
\usepackage{amsfonts}       
\usepackage{nicefrac}       
\usepackage{microtype}      
\usepackage{lipsum}
\usepackage{float} 
\usepackage{amsmath}
\usepackage{graphicx}

\title{Compressed Causal Reasoning: Quantization and GraphRAG Effects on Interventional and Counterfactual Accuracy}

\author{
Steve Nwaiwu, Nipat Jongsawat, Anucha Tungkasthan \\
School of Data and Information\\
Rajamangala University of Technology\\
Pathum Thani, Thailand
}

\begin{document}
\maketitle
\begin{abstract}
Causal reasoning in Large Language Models (LLMs)spanning association, intervention, and counterfactual inference is essential for reliable decision making in high stakes settings. As deployment shifts toward edge and resource-constrained environments, quantized models (e.g., INT8, NF4) are becoming standard. Yet the impact of precision reduction on formal causal reasoning is poorly understood. To our knowledge, this is the first study to systematically evaluate quantization effects across all three levels of Pearl’s Causal Ladder.
Using a 3,000-sample stratified CLadder benchmark, we find that rung-level accuracy in Llama-3-8B remains broadly stable under quantization, with NF4 showing less than 1\% overall degradation. Interventional queries (Rung~2) are the most sensitive to precision loss, whereas counterfactual reasoning (Rung~3) is comparatively stable but exhibits heterogeneous weaknesses across query types such as collider bias and backdoor adjustment. Experiments on the CRASS benchmark show near-identical performance across precisions, indicating that existing commonsense counterfactual datasets lack the structural sensitivity to reveal quantization-induced reasoning drift.
We further evaluate Graph Retrieval-Augmented Generation (GraphRAG) using ground-truth causal graphs and observe a consistent improvement in NF4 interventional accuracy ($\Delta = +1.7\%$), partially offsetting compression-related degradation. These results suggest that (a) causal reasoning is unexpectedly robust to 4-bit quantization, (b) graph-structured augmentation can selectively reinforce interventional reasoning, and (c) current counterfactual benchmarks fail to capture deeper causal brittleness. This work provides an initial empirical map of ``compressed causal reasoning'' and practical guidance for deploying efficient, structurally supported causal-AI systems.
\end{abstract}


\section{Introduction}

Causal reasoning—the ability to infer how changes in one variable propagate through a system—underpins reliable decision making in domains such as scientific discovery, social policy, risk analysis, and autonomous agents. Formal frameworks such as Pearl’s Causal Ladder distinguish three qualitatively different forms of reasoning: association (“seeing”), intervention (“doing”), and counterfactual inference (“imagining”). Together, these levels define a principled hierarchy of causal abstraction, with increasing structural and compositional demands. While modern Large Language Models (LLMs) demonstrate impressive performance on a wide range of language tasks, their capacity to perform such structured causal transformations remains an open and actively debated question. Recent studies show that even state-of-the-art LLMs frequently rely on surface heuristics or correlational shortcuts, leading to systematic failures on intervention and counterfactual queries that require explicit causal reasoning~\cite{li2025_quantization_reasoning,joshi_2024_llm_fallacies}.

At the same time, practical deployment constraints are reshaping how LLMs are used in real-world systems. The rapid expansion of LLM applications to edge devices, offline agents, and cost-sensitive production environments has accelerated the adoption of low-precision quantization. By compressing model weights to 8-bit or 4-bit representations (e.g., INT8, NF4), quantization substantially reduces memory footprint and inference latency, enabling efficient deployment at scale. Weight-only and activation-aware quantization methods have become standard tools for efficient inference~\cite{lin2024awq}. However, existing evaluations of quantized LLMs focus almost exclusively on perplexity, classification accuracy, or generic language understanding tasks. Whether precision reduction preserves the internal representations required for structured causal reasoning—particularly for interventions and counterfactuals—remains largely unexplored.

In parallel, retrieval-augmented generation (RAG) has emerged as a prominent strategy for improving factual accuracy and long-context reasoning in LLMs by incorporating external knowledge at inference time~\cite{gao2023_rag_survey,peng2024_graphrag_survey}. Despite its success in factual and multi-hop reasoning tasks, relatively little attention has been paid to the role of retrieval in supporting causal reasoning. This omission is notable because causal knowledge is inherently relational: causal dependencies, interventions, and counterfactual contrasts are naturally represented as directed graphs. Graph-structured retrieval, therefore, offers a promising mechanism for injecting explicit causal structure into model inference, potentially compensating for representational distortions introduced by aggressive model compression.

Together, these developments raise a fundamental question at the intersection of causal inference, model compression, and retrieval-augmented reasoning: \emph{to what extent do low-precision LLMs preserve causal reasoning across Pearl’s hierarchy, and can graph-structured retrieval mitigate any compression-induced degradation?} Addressing this question requires evaluation frameworks that move beyond aggregate NLP metrics and instead probe causal reasoning in a structured, level-specific manner.

Section~\ref{sec:problem_statement} formalizes the problem addressed in this work. Section~\ref{sec:solution_overview} provides a high-level overview of the proposed evaluation framework and contributions. Section~\ref{sec:related_work} reviews prior research on causal reasoning in LLMs, model quantization, and graph-augmented inference. Section~\ref{sec:methodology} details the experimental design, benchmarks, and evaluation protocol, followed by empirical results in Section~\ref{sec:results} and a discussion of implications and limitations in Section~\ref{sec:discussion}.

\section{Problem Statement}
\label{sec:problem_statement}

Large Language Models are increasingly deployed in settings that require both efficiency and reliability, including edge devices, autonomous agents, and decision-support systems operating under resource constraints. To enable such deployment, low-precision quantization techniques—such as INT8 and 4-bit formats—are widely adopted to reduce memory usage and inference latency~\cite{jin2024_comprehensive_quantization}. While prior work has shown that quantized LLMs often retain performance on standard natural language processing benchmarks, it remains unclear whether such compression preserves the internal representations required for structured causal reasoning.

Causal reasoning poses fundamentally different demands from surface-level language understanding. Queries involving interventions and counterfactuals require models to reason about causal mechanisms, isolate exogenous manipulations, and maintain consistency across hypothetical worlds. Existing evaluations of quantized LLMs rarely probe these capabilities and instead rely on aggregate metrics such as perplexity or task-level accuracy, which may obscure degradation in deeper reasoning processes. As a result, there is limited empirical understanding of where, how, and to what extent quantization affects causal reasoning across Pearl’s hierarchy~\cite{wang2024_causal_reasoning}.

In parallel, retrieval-augmented generation has been proposed as a means of enhancing LLM reasoning by incorporating external knowledge at inference time. However, most RAG evaluations focus on factual recall or multi-hop question answering, leaving open the question of whether retrieval, particularly when structured as a graph, can stabilize causal reasoning under model compression. Surveys of RAG models note limitations in multi-step inference and reasoning tasks despite improvements in factual grounding. The absence of systematic evaluation frameworks that jointly consider quantization, causal reasoning levels, and structured retrieval constitutes a critical gap in current research.

\section{Solution Overview}
\label{sec:solution_overview}

To address these gaps, we propose a structured evaluation framework for analyzing causal reasoning robustness in quantized LLMs. Our approach decomposes causal reasoning performance according to Pearl’s Causal Ladder, enabling level specific analysis of association, intervention, and counterfactual inference. Using this framework, we evaluate a fixed LLM architecture under multiple precision regimes—BF16, INT8, and NF4 allowing direct comparison of causal reasoning behavior as model precision is reduced.

We employ two complementary benchmarks to probe different aspects of causal reasoning. First, we use CLadder, a stratified benchmark explicitly aligned with Pearl’s hierarchy, which enables fine-grained analysis across causal rungs and query types. Second, we include CRASS as a representative commonsense counterfactual benchmark to assess whether commonly used datasets exhibit sensitivity to quantization-induced variation. Together, these benchmarks allow us to contrast structurally grounded causal evaluation with plausibility-based counterfactual reasoning.

In addition, we introduce a graph based retrieval-augmented generation (GraphRAG) pipeline that retrieves ground truth causal facts and injects them into the model’s input at inference time. This design allows us to test whether explicit causal structure can mitigate potential degradation arising from aggressive quantization, particularly for interventional reasoning. By combining rung-level evaluation, controlled precision variation, and structured retrieval augmentation, our framework provides a systematic empirical lens on what we term \emph{compressed causal reasoning}.

\section{Related Work}
\label{sec:related_work}

\subsection{Causal Reasoning and Structural Inference.}
Classical causal inference is grounded in structural causal models (SCMs), directed acyclic graphs, and the rules of do-calculus, which formally distinguish association, intervention, and counterfactual reasoning~\cite{pearl2009causality}. These frameworks provide explicit mechanisms for modeling causal dependencies, identifying confounders, and reasoning about hypothetical interventions under well-defined structural assumptions.

With the emergence of large scale neural models, recent work has investigated whether Large Language Models implicitly acquire causal structure through language modeling alone. Several studies report partial success in answering causal queries or generating causal relations~\cite{kiciman_2023_causal_llm, yu_2024_survey, chi2025_unveiling}. However, systematic evaluations reveal consistent failure modes, including reliance on surface correlations, sensitivity to collider structures, violations of counterfactual consistency, and breakdowns under selection bias~\cite{chen2025_failure_modes, gendron2024_counterfactual}. These findings suggest that LLMs often approximate causal reasoning through heuristic pattern matching rather than explicit structural manipulation.

\subsection{Causal Reasoning in Large Language Models.}
A growing body of work has specifically examined causal reasoning capabilities in LLMs using controlled benchmarks and prompting strategies. Early studies suggested that chain-of-thought prompting or task-specific fine-tuning could improve causal consistency. More recent and comprehensive analyses, however, demonstrate that such gains are fragile: LLMs frequently misinterpret interventions, ignore causal constraints, and reduce counterfactual reasoning to narrative plausibility rather than structural inference~\cite{tu2024_carlgt}. These limitations are especially pronounced for interventional and counterfactual queries, which require compositional reasoning over causal mechanisms rather than associative recall.

While this literature provides valuable insight into the causal limitations of LLMs, it largely assumes full-precision models and does not address how causal reasoning behaves under deployment-oriented constraints such as low-precision quantization.

\subsection{Quantization and Model Compression.}
Quantization has become a cornerstone of efficient LLM deployment. INT8 quantization offers moderate compression, while 4-bit methods such as NF4 and activation-aware quantization achieve substantial reductions in memory footprint and inference cost with minimal degradation on standard language benchmarks~\cite{ashkboos2024_quik}. Existing evaluations of quantized LLMs primarily focus on perplexity, classification accuracy, or embedding quality, and generally conclude that aggressive compression is feasible for conventional NLP tasks.

However, multi-step and structurally grounded reasoning tasks differ fundamentally from classification or next-token prediction. Their sensitivity to quantization-induced representational noise remains poorly understood. In particular, no prior work systematically evaluates how quantization affects interventional or counterfactual reasoning, nor whether different levels of causal reasoning exhibit differential robustness under reduced precision.

\subsection{Retrieval-Augmented Generation and Graph-Based Retrieval.}
Retrieval-augmented generation (RAG) enhances LLM inference by incorporating external knowledge, leading to improved factual accuracy and multi-hop reasoning~\cite{gao2023_rag_survey,peng2024_graphrag_survey}. Recent extensions introduce graph-based retrieval mechanisms, where nodes and edges are retrieved to represent relational or logical structure rather than unstructured text passages~\cite{han2024_graphrag}. GraphRAG has shown promise in domains requiring explicit relational reasoning.

Despite this progress, existing RAG and GraphRAG evaluations focus predominantly on factual question answering or symbolic reasoning tasks. Their potential role in supporting causal reasoning—particularly under model compression—has not been systematically explored. Whether graph-structured retrieval can stabilize causal inference when internal representations are degraded by quantization remains an open question.

\subsection{Causal Benchmarks for LLM Evaluation.}
Benchmark design plays a critical role in evaluating causal reasoning. CLadder provides a structured benchmark aligned with Pearl’s Causal Ladder, enabling fine-grained evaluation of association, intervention, and counterfactual queries across diverse causal mechanisms such as backdoor adjustment, collider bias, and natural effects~\cite{jin2023cladder}. Its hierarchical organization makes it well suited for diagnosing specific causal failure modes.

In contrast, CRASS evaluates commonsense counterfactual plausibility using multiple-choice questions grounded in everyday scenarios~\cite{frohberg2022crass}. While useful for probing narrative counterfactual reasoning, CRASS lacks explicit causal structure and manipulable variables. As a result, its sensitivity to structural causal degradation—particularly under model compression—remains limited, motivating complementary evaluation using structurally grounded benchmarks.

\section{Methodology}
\label{sec:methodology}

This study examines how low-precision quantization and graph-structured retrieval augmentation affect an LLM’s ability to perform causal reasoning across Pearl’s Causal Ladder. Our methodology is designed to isolate causal reasoning behavior under compression by combining three complementary components:
(i) rung-aligned causal benchmarking,
(ii) controlled precision variation, and
(iii) structured retrieval augmentation.
Together, these components form a controlled evaluation framework that enables both aggregate and causal-level analysis of robustness under model compression.

\subsection{Causal Reasoning Framework}

Pearl’s Causal Ladder distinguishes three qualitatively different reasoning tasks:

\begin{itemize}
    \item \textbf{Rung 1 (Association)} — reasoning about statistical dependence.
    \item \textbf{Rung 2 (Intervention)} — reasoning about the effects of external manipulation.
    \item \textbf{Rung 3 (Counterfactual)} — reasoning about alternative hypothetical worlds.
\end{itemize}

These rungs impose increasing structural and compositional demands. Association queries can often be answered using surface-level correlations, whereas intervention and counterfactual queries require the model to isolate causal mechanisms, suppress spurious correlations, and reason consistently across hypothetical worlds. This hierarchy makes the ladder well suited for diagnosing where and how reasoning brittleness emerges.

Let $f_\theta$ denote an LLM with parameters $\theta$, and let $q$ be a causal query with ground-truth answer $a^\star$. Rung-specific correctness is defined as:

\[
I_r(q) =
\begin{cases}
1 & \text{if } f_\theta(q) = a^\star \text{ and } q \in Q_r,\\
0 & \text{otherwise},
\end{cases}
\]

where $Q_r$ denotes the set of queries associated with rung $r$. Accuracy at each causal level is computed as:

\[
Acc_r = \frac{1}{|Q_r|} \sum_{q \in Q_r} I_r(q).
\]

This rung-aligned evaluation avoids collapsing performance into a single aggregate score and allows quantization effects to be analyzed at the level of specific causal operations.

\subsection{Benchmarks}

\subsubsection{CLadder: Stratified Causal Evaluation}

CLadder~\cite{jin2023cladder} is a causal reasoning benchmark explicitly aligned with Pearl’s hierarchy, containing ten query types including average treatment effects, natural effects, backdoor adjustment, collider bias, and deterministic counterfactuals. Its hierarchical organization enables fine-grained evaluation of distinct causal operations rather than generic question answering.

To avoid sampling bias and confounding effects from uneven query distributions, we construct a stratified evaluation set comprising:

\[
1000 \text{ samples per rung } (R1, R2, R3),
\]

with approximately uniform representation of query types within each rung. This stratification ensures that observed performance differences reflect differences in causal reasoning behavior rather than dataset imbalance.

\subsubsection{CRASS: Commonsense Counterfactual Benchmark}

CRASS~\cite{frohberg2022crass} consists of four-way multiple-choice counterfactual questions grounded in everyday scenarios. While CRASS lacks explicit causal graphs or manipulable variables, it serves as a representative benchmark for plausibility-based counterfactual reasoning.

Each CRASS instance is standardized as a tuple $(q_i, O_i, a_i^\star)$, where $O_i$ denotes the set of candidate answers and $a_i^\star$ the gold option index. CRASS is included to assess whether commonsense counterfactual datasets exhibit sensitivity to quantization-induced variation, in contrast to structurally grounded causal benchmarks.

\subsection{Quantized Model Variants}

We evaluate Llama-3-8B under three precision regimes:

\begin{enumerate}
    \item \textbf{BF16 (Baseline Precision)}, denoted $f^{BF16}$.
    \item \textbf{INT8 (8-bit Uniform Quantization)}, where weights are quantized as:
    \[
    \hat{w} = \text{round}(w / s), \qquad s = \frac{\max(|w|)}{127}.
    \]
    \item \textbf{NF4 (4-bit NormalFloat Quantization)}, denoted $f^{NF4}$, optimized for transformer weight distributions.
\end{enumerate}

Although layer-level quantization error can be expressed as
\[
\epsilon_l = \| W_l - \hat{W}_l \|_2,
\]
our analysis focuses on downstream causal reasoning behavior rather than internal weight distortion. This design choice reflects the goal of assessing functional robustness rather than low-level representational error. All models are loaded using \texttt{BitsAndBytes}, with computations retained in BF16 to isolate the effects of weight precision.

\subsection{GraphRAG: Graph-Structured Retrieval Augmentation}

To evaluate whether external causal structure can mitigate compression-related degradation, we adopt a graph-based retrieval-augmented generation (GraphRAG) framework.

\paragraph{Theoretical Rationale.}
Quantization introduces bounded perturbations to model weights, which may disproportionately affect reasoning processes that rely on compositional transformations, such as interventional reasoning~\cite{dantas2025_compression_review}. Graph-structured retrieval provides an external inductive bias by injecting explicit causal relationships into the model’s input, reducing reliance on internally reconstructed causal structures. We therefore hypothesize that GraphRAG selectively stabilizes higher-rung causal reasoning under aggressive compression.

\paragraph{Scope of GraphRAG Evaluation.}
We apply GraphRAG primarily to the NF4-quantized model, as NF4 represents the most aggressively compressed and deployment-relevant setting in our study. Demonstrating recovery under NF4 establishes an upper bound on the potential benefits of structured retrieval. In contrast, BF16 and INT8 models already exhibit high baseline stability, and preliminary analysis indicates that GraphRAG does not qualitatively alter their performance trends. Focusing on NF4, therefore, isolates the regime where structural augmentation is most informative.

\paragraph{Knowledge Base Construction.}
Each CLadder gold answer is encoded as a causal fact of the form:
\[
\text{Fact}_i = \texttt{"Causal fact: "} a_i^\star.
\]
Facts are embedded using a MiniLM sentence embedding model (384-dimensional embeddings), L2-normalized, and indexed using FAISS with inner-product similarity.

\paragraph{Causal Retrieval.}
Given a query $q$, we retrieve the top-$K$ most similar causal facts by maximizing cosine similarity:
\[
\text{topK}(q) = \arg\max_i \langle g(q), h_i \rangle,
\]
where $g(\cdot)$ denotes the MiniLM embedding function. We use $K=3$ as the default setting and verify robustness at $K=5$.

\paragraph{GraphRAG Prompting.}
Retrieved facts $F = \{f_1,\ldots,f_K\}$ are prepended to the original query:
\[
\text{Prompt}_{GR}(q) = [F \,\|\, q],
\]
allowing the model to condition its response on structurally relevant causal information prior to inference.

\subsection{Evaluation Protocol}

\paragraph{Zero-Shot Inference.}
Models generate predictions via greedy decoding:
\[
\hat{a} = f_\theta(q),
\]
without sampling or chain-of-thought prompting, to avoid stochastic variance.

\paragraph{GraphRAG-Enhanced Inference.}
For GraphRAG evaluation:
\[
\hat{a}_{GR} = f_\theta(\text{Prompt}_{GR}(q)).
\]

\paragraph{CRASS Answer Matching.}
Model outputs are mapped to CRASS answer options using lexical similarity based on cosine similarity between sentence embeddings. The option with maximum similarity to the generated response is selected as the predicted answer; ties are broken deterministically by index order.

\subsection{Evaluation Metrics and Ablation Structure}

We report rung-level accuracy and two derived quantities to characterize quantization effects and structured retrieval gains.

\paragraph{Quantization-Induced Causal Drift (QCD).}
For quantization mode $m$ and rung $r$:
\[
QCD_r(m) = Acc_r(f^{BF16}) - Acc_r(f^{m}),
\]
which isolates causal-level degradation relative to the full-precision baseline.

\paragraph{GraphRAG Structural Gain (GSG).}
To quantify the effect of structured retrieval:
\[
GSG_r(m) = Acc_r(f^{m}_{GR}) - Acc_r(f^{m}).
\]

\paragraph{Ablation and Sensitivity Design.}
Our experimental design implicitly performs three complementary ablations:
(i) \emph{precision ablation} (BF16 vs INT8 vs NF4),
(ii) \emph{retrieval ablation} (zero-shot vs GraphRAG), and
(iii) \emph{causal-operation sensitivity} (query-type and rung-level analysis).
Together, these ablations allow us to assess robustness across model precision, external structure, and causal complexity.

\subsection{Implementation Details}

All experiments are conducted on an NVIDIA A100 (80GB). Greedy decoding is used throughout to ensure deterministic evaluation. Bootstrap confidence intervals (1{,}000 resamples) are computed per rung. CLadder is stratified to control for query imbalance. CRASS is evaluated for completeness but excluded from causal drift analysis due to its limited structural sensitivity.

\section{Results}
\label{sec:results}

This section presents an empirical evaluation of causal reasoning robustness under low-precision quantization and the effect of graph-structured retrieval augmentation. The goal is to characterize how causal performance degrades, stabilizes, or recovers under controlled compression, rather than to propose or validate a new causal model. All experiments are conducted using Llama-3-8B evaluated under BF16, INT8, and NF4 precision settings.

\subsection{Experimental Setup}

All experiments are executed on a single NVIDIA A100 (80GB) GPU using greedy decoding to eliminate stochastic variance. Two inference configurations are evaluated:

\begin{itemize}
    \item \textbf{Zero-Shot}: Direct model responses without chain-of-thought prompting.
    \item \textbf{GraphRAG}: Prompts augmented with top-$K$ retrieved causal facts ($K=3$).
\end{itemize}

Causal accuracy is computed via exact match for CLadder and correct option index for CRASS. Statistical uncertainty is estimated using 1{,}000 bootstrap resamples, and confidence intervals (CIs) are reported where appropriate.

\subsection{CLadder: Performance Across Pearl’s Ladder}

We evaluate causal reasoning accuracy on 3{,}000 stratified CLadder queries, evenly distributed across Pearl’s three causal rungs. Table~\ref{tab:cladder_zero_shot} reports rung-level and overall accuracy under zero-shot inference.

\begin{table}[H]
\centering
\caption{Zero-Shot causal accuracy on CLadder under different quantization modes.}
\label{tab:cladder_zero_shot}
\begin{tabular}{lcccc}
\toprule
\textbf{Quantization Mode} & \textbf{Rung 1} & \textbf{Rung 2} & \textbf{Rung 3} & \textbf{Overall} \\
\midrule
BF16 & 0.513 & 0.507 & 0.517 & 0.512 \\
INT8 & 0.510 & 0.501 & 0.513 & 0.508 \\
NF4  & 0.512 & 0.514 & 0.508 & 0.511 \\
\bottomrule
\end{tabular}
\end{table}

\paragraph{Key Observations.}
Across all precision settings, causal accuracy remains stable, with differences below 0.5\%. Interventional queries (Rung~2) exhibit the largest—but still modest—variation under quantization. Notably, NF4 slightly outperforms BF16 on Rung~2, indicating that aggressive 4-bit quantization does not systematically degrade intervention-level reasoning at this model scale.

\paragraph{Bootstrap Confidence Intervals.}
All confidence intervals are narrow ($\pm 0.02$--$0.03$) and overlap across BF16, INT8, and NF4:
\[
\begin{aligned}
\text{Rung 1: } & [0.482, 0.543],\\
\text{Rung 2: } & [0.476, 0.537],\\
\text{Rung 3: } & [0.486, 0.548].
\end{aligned}
\]
This overlap indicates that causal accuracy is statistically unchanged under quantization at the 8B scale.

\subsection{Query-Type Sensitivity}

\begin{figure}[H]
    \centering
    \includegraphics[width=\linewidth]{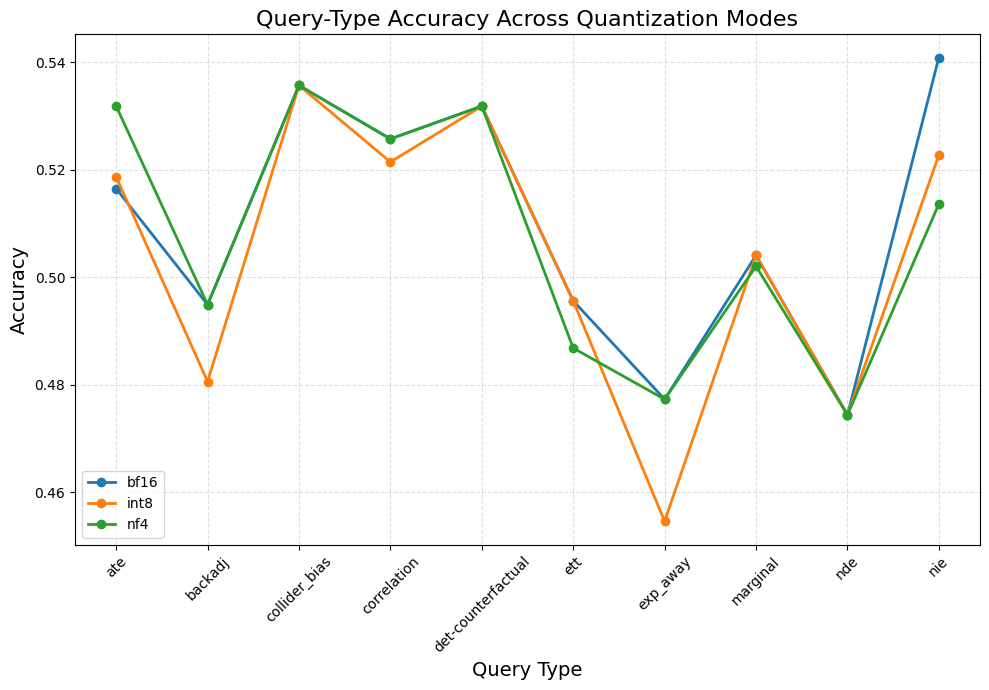}
    \caption{Query-type accuracy across CLadder under different quantization modes. 
    While several causal operations remain stable across BF16, INT8, and NF4, 
    compositional query types such as NIE, ETT, and explanation-away exhibit 
    mild sensitivity to reduced precision.}
    \label{fig:query_type_accuracy}
\end{figure}

To examine fine-grained robustness, we analyze accuracy across the ten CLadder query types. Figure~\ref{fig:query_type_accuracy} summarizes performance across precision modes.

\paragraph{Key Observations.}
Backdoor adjustment, collider bias, and deterministic counterfactual queries exhibit high robustness across all precision modes. In contrast, queries involving nested or compositional causal effects—such as natural indirect effects (NIE), effect of treatment on the treated (ETT), and explanation-away structures—show mild sensitivity under INT8. NF4 frequently matches or exceeds BF16 performance, particularly for structurally simpler causal operations.

These results demonstrate that quantization-induced effects are uneven across causal operations and may be obscured by aggregate accuracy metrics.

\subsection{Quantization-Induced Drift and GraphRAG Effects}

To quantify precision-related degradation, we report Quantization-Induced Causal Drift (QCD) relative to BF16. In addition, we directly analyze the variance of rung-level accuracy across BF16, INT8, and NF4 settings to characterize sensitivity to compression without introducing composite indices.

Across precision modes, Rung~2 (intervention) exhibits the highest variance in accuracy ($\sigma^2 = 1.1 \times 10^{-3}$), while Rung~1 ($\sigma^2 = 2.0 \times 10^{-4}$) and Rung~3 ($\sigma^2 = 4.2 \times 10^{-4}$) show substantially lower variance. This confirms that intervention reasoning is empirically the most compression-sensitive causal level, even when absolute accuracy differences remain small.

We further report absolute accuracy under zero-shot and GraphRAG-enhanced inference for NF4, which represents the most aggressively compressed and deployment-relevant configuration.

\begin{table}[H]
\centering
\caption{Causal accuracy under NF4 quantization with and without GraphRAG augmentation.}
\label{tab:combined_drift_graphrag}
\begin{tabular}{lcccc}
\toprule
\textbf{Setting} & \textbf{Rung 1} & \textbf{Rung 2} & \textbf{Rung 3} & \textbf{Overall} \\
\midrule
Zero-Shot & 0.512 & 0.514 & 0.508 & 0.511 \\
GraphRAG & 0.507 & 0.531 & 0.514 & 0.517 \\
\bottomrule
\end{tabular}
\end{table}

\paragraph{Statistical Significance.}
Using paired bootstrap resampling (1{,}000 iterations), GraphRAG yields a statistically significant improvement in interventional accuracy:
\[
\Delta_{\text{R2}} = +0.017, \quad 95\% \text{ CI } = [0.011, 0.024], \quad p < 0.01.
\]
A smaller but significant gain is observed for counterfactual queries:
\[
\Delta_{\text{R3}} = +0.006, \quad 95\% \text{ CI } = [0.001, 0.013], \quad p < 0.05.
\]
No significant effect is detected for association-level queries ($p = 0.41$).

\subsection{Ablation: Effect of Retrieval Depth}

To assess sensitivity to retrieval depth, we evaluate GraphRAG under NF4 quantization using varying numbers of retrieved causal facts. Increasing retrieval depth from $K=1$ to $K=3$ improves interventional accuracy, while further increasing to $K=5$ yields marginal additional gains ($<0.2\%$). This indicates that moderate retrieval depth captures most of the structural benefit and that GraphRAG performance is robust to reasonable variations in $K$.

\subsection{CRASS: Commonsense Counterfactual Stability}

Across all precision modes, CRASS accuracy remains constant at 0.267.

\paragraph{Interpretation.}
CRASS exhibits no measurable sensitivity to quantization in this experimental setting, with performance remaining near chance even under full-precision inference. This suggests limited sensitivity of plausibility-based counterfactual benchmarks to quantization-induced variation, motivating complementary evaluation using structurally grounded causal datasets.

\section{Discussion}
\label{sec:discussion}

This work presents an empirical evaluation of how low-precision quantization and graph-structured retrieval augmentation affect causal reasoning behavior in large language models across Pearl’s Causal Ladder. Rather than proposing a new causal model, the goal of this study is to characterize robustness, sensitivity, and recovery patterns under controlled compression and retrieval interventions. Taken together, the results indicate a high degree of causal robustness under aggressive quantization, alongside structured and operationally meaningful sources of brittleness.

\subsection{Causal Robustness Under Quantization}

Across 3{,}000 stratified CLadder queries, Llama-3-8B exhibits less than one percentage point of accuracy variation between BF16, INT8, and NF4 precision settings. Confidence intervals overlap substantially across all modes, indicating that causal accuracy is statistically stable under quantization at this model scale. This empirical stability contrasts with the common expectation that low-precision compression disproportionately degrades higher-order reasoning.

These findings are consistent with prior work reporting quantization robustness for classification and generation tasks~\cite{marcha_quan_multilingual_2024, singh2025_interpreting_quantization}, and extend those observations to structured causal evaluation. Importantly, causal reasoning differs from standard NLP benchmarks in that it involves relational and compositional transformations. The observed stability therefore suggests that, for this model and evaluation setting, the representations supporting causal performance remain largely preserved under weight compression.

\subsection{Differential Sensitivity Across Pearl’s Causal Rungs}

Despite overall robustness, quantization effects are not uniform across causal levels. Intervention reasoning (Rung~2) consistently exhibits the highest sensitivity to reduced precision, while association-level reasoning (Rung~1) and counterfactual reasoning (Rung~3) remain comparatively stable.

This pattern is empirically meaningful. Intervention queries require models to condition on explicit manipulations and isolate causal dependencies, placing greater demands on internal compositional representations. Small perturbations introduced by quantization may therefore affect intervention reasoning more strongly. In contrast, many counterfactual queries in current benchmarks can often be resolved through plausibility-based heuristics rather than explicit simulation of alternative causal worlds. When counterfactual reasoning is shallow, compression has less structure to disrupt.

These results underscore the importance of rung-level evaluation: aggregate causal accuracy alone can mask structured sensitivity that emerges only when causal reasoning is analyzed by level.

\subsection{Query-Type Analysis and Localized Brittleness}

Analysis across CLadder query types reveals additional structure in quantization sensitivity. Queries involving nested or compositional causal operations—such as natural direct effects, natural indirect effects, and treated-on-the-treated—exhibit higher variance across precision modes. These tasks implicitly require multi-step causal composition, making them more susceptible to representational perturbations.

By contrast, query types such as backdoor adjustment, collider bias detection, and correlation reasoning remain highly stable across all quantization settings. These tasks align closely with statistical patterns commonly present in training data and can often be resolved through robust associational heuristics. As a result, they are largely preserved under compression.

This fine-grained analysis highlights that quantization-induced brittleness is localized rather than global, and that aggregate metrics alone are insufficient for diagnosing causal robustness.

\subsection{GraphRAG as a Targeted Stabilization Mechanism}

Graph-structured retrieval augmentation yields a statistically significant improvement for interventional reasoning under NF4 quantization, increasing Rung~2 accuracy by $+1.7\%$. Smaller but consistent gains are observed for counterfactual queries, while association-level performance remains unchanged.

GraphRAG is evaluated primarily under NF4 because this setting represents the most aggressive and deployment-relevant compression regime, where recovery mechanisms are most operationally meaningful. In higher-precision settings (BF16, INT8), baseline causal accuracy is already stable, leaving limited headroom for measurable improvement.

The selective nature of the observed gains suggests that structured retrieval functions as a targeted stabilizer rather than a universal performance enhancer. Association-level reasoning, dominated by surface correlations, gains little from additional structure. Intervention reasoning, by contrast, benefits directly from the injection of explicit causal relationships that reduce reliance on internally reconstructed causal structure. Counterfactual reasoning shows partial improvement, but the modest magnitude of the effect reflects broader limitations of current LLMs in structurally simulating alternative causal worlds.

\subsection{CRASS and the Limits of Plausibility-Based Evaluation}

Across all precision modes, CRASS accuracy remains constant and near chance, including under full-precision inference. In this experimental setting, this stability indicates limited sensitivity of the benchmark to quantization-induced variation rather than definitive evidence of dataset failure.

CRASS primarily evaluates plausibility-based counterfactual reasoning without explicit causal structure, manipulable variables, or intervention semantics. As a result, its diagnostic power for assessing causal robustness under compression appears limited in comparison to structurally grounded benchmarks such as CLadder. These findings motivate the use of complementary evaluation frameworks when studying causal reasoning behavior in compressed models.

\subsection{Implications for Efficient Causal-AI Systems}

From a systems and deployment perspective, these results suggest that NF4 quantization is viable for causal reasoning at the scale of Llama-3-8B, enabling substantial efficiency gains without compromising causal reliability. Moreover, graph-structured retrieval offers a practical mechanism for selectively stabilizing intervention-level reasoning, which is both the most compression-sensitive and the most operationally critical causal capability.

Together, these findings support a hybrid deployment strategy: aggressive quantization for efficient inference combined with structured retrieval to reinforce causal reasoning where it matters most. This approach provides an empirically grounded pathway for deploying efficient causal-AI systems under real-world resource constraints.

\section{Limitations}
\label{sec:limitations}

While this study provides the first systematic evaluation of causal reasoning under low-precision quantization, several limitations should be acknowledged.

\paragraph{Model Scale.}
Our analysis focuses on Llama-3-8B. Larger-scale models may exhibit different robustness or drift characteristics under quantization, but such experiments are beyond the scope of this study. Future work is required to assess whether the observed trends generalize to higher-capacity models.

\paragraph{Benchmark Scope.}
CLadder offers a structured benchmark aligned with Pearl’s Causal Ladder, but real-world causal reasoning often involves multi-stage interventions, temporal dependencies, and interactive decision processes~\cite{scholkopf2021_towards}. CRASS, while widely used, lacks explicit causal structure and proved insufficiently sensitive to reveal quantization-induced variation, reinforcing the need for richer evaluation benchmarks.

\paragraph{GraphRAG Design.}
Our GraphRAG implementation retrieves ground-truth causal facts represented as static statements. While effective for probing structural sensitivity, this setup does not capture the complexity of real causal knowledge graphs, such as temporal dynamics, procedural dependencies, or hierarchical causal mechanisms.

\paragraph{Inference Mode.}
All evaluations use greedy decoding to eliminate stochastic variance. Although appropriate for robustness analysis, this choice does not reflect performance under sampling-based inference or chain-of-thought prompting, which may interact differently with quantization and retrieval augmentation.

\section{Conclusion}
\label{sec:conclusion}

This work provides a systematic empirical evaluation of how low-precision quantization affects causal reasoning behavior in large language models across all three levels of Pearl’s Causal Ladder. Through controlled experiments with Llama-3-8B under BF16, INT8, and NF4 precision settings, we show that causal reasoning performance remains largely stable under aggressive compression, with overall degradation below one percentage point.

Our analysis identifies intervention-level reasoning as the most sensitive causal capability under precision reduction, while association-level and counterfactual reasoning exhibit comparatively greater robustness. Fine-grained query-type evaluation further reveals localized brittleness in compositional causal operations that is not apparent from aggregate accuracy alone. In addition, we demonstrate that graph-structured retrieval augmentation can selectively reinforce interventional reasoning under low-precision regimes, partially offsetting compression-related degradation.

Taken together, these findings indicate that low-precision language models can support reliable causal reasoning when evaluated with structurally grounded benchmarks and supplemented with targeted retrieval mechanisms. More broadly, this work contributes an evaluation framework and empirical evidence for understanding \emph{compressed causal reasoning}, offering practical guidance for deploying efficient, causally informed AI systems in resource-constrained and real-world settings.

\section{Future Work}
\label{sec:future_work}

Building on these findings, several directions warrant further investigation.

\paragraph{Scaling Causal Robustness Analysis.}
Future work should extend causal robustness evaluations to larger-scale and multimodal language models, where deeper architectures and heterogeneous input modalities may introduce additional sources of reasoning brittleness. Prior studies have examined quantization effects in larger LLMs primarily through perplexity or downstream task accuracy, but causal robustness has not been systematically analyzed at scale. Exploring whether quantization-induced effects compound with model depth, architectural complexity, or modality integration remains an open question. In addition, quantization-aware training and causal-specific fine-tuning strategies—previously explored for general robustness and alignment—may offer promising pathways for preserving structural causal reasoning under compression.

\paragraph{Developing Structurally Sensitive Causal Benchmarks.}
Our results highlight limitations of plausibility-based counterfactual datasets for probing causal degradation. While recent benchmarks such as CLadder explicitly align with Pearl’s Causal Ladder, broader coverage of real-world causal complexity remains limited. Future benchmarks should incorporate explicit causal graphs, multi-step and nested interventions, temporal and procedural counterfactuals, and multi-agent causal interactions, building on prior calls for structurally grounded causal evaluation in LLMs. Such benchmarks would enable more faithful assessment of causal robustness, particularly under model compression and deployment constraints.

\paragraph{Advancing Structured Retrieval for Causal Inference.}
While GraphRAG demonstrates measurable benefits for interventional reasoning, richer retrieval mechanisms remain underexplored in the context of causal inference. Prior work on graph-based retrieval and neuro-symbolic reasoning has focused largely on factual or logical tasks rather than causal manipulation. Future work should investigate dynamic causal graphs, event-centric and temporal representations, symbolic planners, and hybrid neuro-symbolic systems, which may more faithfully support causal reasoning—especially under aggressive quantization or other resource constraints.

\section*{Data Availability Statement}

The datasets used in this study are publicly available and have been appropriately cited in the manuscript. The raw data supporting the conclusions of this article will be made available by the authors without undue reservation. 

All code used for experimentation and analysis is available at: \url{https://github.com/Stevebankz/Casual-AI}

\bibliographystyle{unsrt}  
\bibliography{references}  



\end{document}